\documentclass{article}

\usepackage[preprint]{neurips_2026}

\usepackage{graphicx}
\usepackage{amsmath}
\usepackage{hyperref} 
\usepackage{cleveref} 
\usepackage{multirow}

\usepackage[ruled]{algorithm2e} 

\SetAlFnt{\small}
\SetAlCapFnt{\small}
\SetAlCapNameFnt{\small}
\SetAlCapHSkip{0pt}

\usepackage[utf8]{inputenc} 
\usepackage[T1]{fontenc}    
\usepackage{url}            
\usepackage{booktabs}       
\usepackage{amsfonts}       
\usepackage{nicefrac}       
\usepackage{microtype}      
\usepackage{xcolor}         

\title{PermaVid: Consistent Video Generation Across Edits via Disentangled Context Memory}

\author{
Shuai Yang*$^{1}$, 
Bingjie Gao*$^{1}$, 
Ziwei Liu$^{3}$,
Jiaqi Wang$^{5}$, \\
\textbf{Dahua Lin$^{4}$,
Tong Wu$^{2}$}
\\
{\small
$^{1}$ Shanghai Jiao Tong University\quad
$^{2}$ Stanford University \quad
$^{3}$ S-Lab, Nanyang Technological University \quad
}
\\
{\small 
$^{4}$ The Chinese University of Hong Kong\quad
$^{5}$ Shanghai Innovation Institute\quad
}\\ \\
\tt\small
        \href{ys-imtech.github.io/projects/PermaVid}{ys-imtech.github.io/projects/PermaVid}
}

\begin{document}

\maketitle
\begin{figure}[htbp]
  \centering
  \maketitle
  \includegraphics[width=1.0\linewidth]{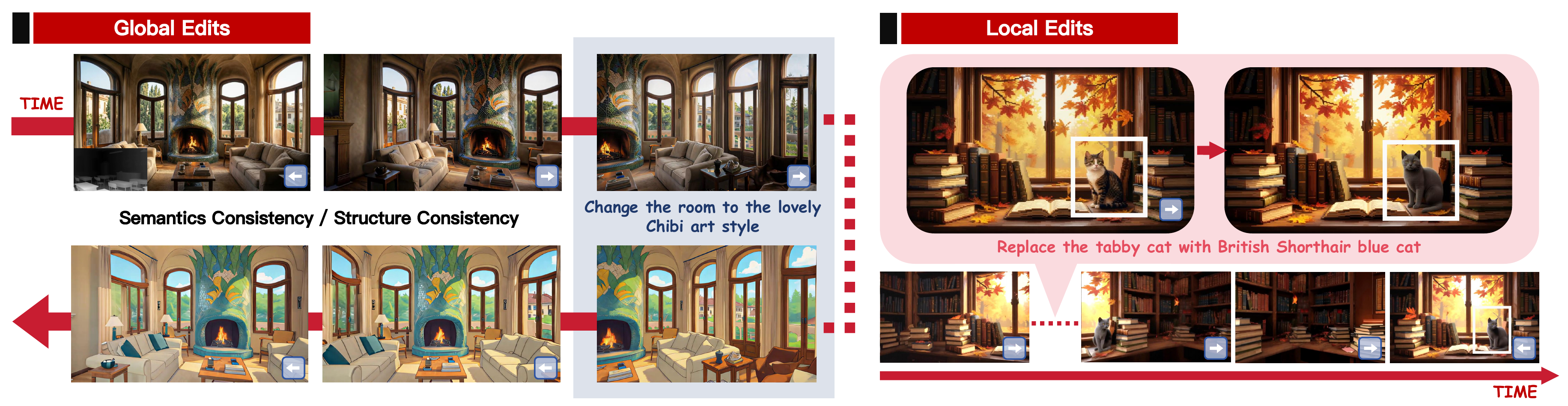}
    \caption{We propose \textbf{PermaVid}, a framework for consistent video generation across edits. For global edits (e.g., style transformation), PermaVid propagates updated semantics consistently across time and viewpoints while maintaining stable geometry. For local edits (e.g., object-level editing), the model reliably recalls the post-edit content during revisiting, preserving both structural integrity and updated local semantics. } 
  \label{fig:teaser}
\end{figure}

\begin{abstract}

Consistent video generation under editing operations requires persistence: when edits modify scene appearance or layout, subsequent generations should remain coherent across time and viewpoints. However, existing memory designs struggle to maintain long-term consistency after such modifications, as stored contexts may become outdated or invalid. To address this, we propose PermaVid, a novel framework built upon a multi-modal context memory that disentangles spatial context into semantic appearance and geometric structure, together with an edit-aware memory update and retrieval strategy that keeps memory evolution aligned with subsequent observations. Specifically, we develop two complementary memory banks: an RGB context memory that captures appearance-aware observations while implicitly encoding geometry, and a depth context memory that preserves geometry-only structure disentangled from semantics. Building on this design, we introduce a memory-guided video generation model that performs multi-modal feature fusion under reference conditions drawn from mixed-modality memory contexts. Experiments demonstrate that our method maintains strong long-term semantic and structural consistency after edits, significantly outperforming state-of-the-art methods.

\end{abstract}

\section{Introduction}
\label{sec:intro}
\vspace{-10pt}

Recent advances in camera-controlled video generation~\cite{videoworldsimulators2024, hong2025relic, sun2025worldplay, wu2025video} and video editing methods~\cite{wan2025wan, yang2024cogvideox, kong2024hunyuanvideo, tan2024imagine360,hacohen2024ltx} have significantly expanded the flexibility of visual content creation. Users can now synthesize videos by specifying camera trajectories, or modify existing videos through editing instructions such as style transformation, object insertion, or scene manipulation. These capabilities enable richer interactive experiences and more controllable generation pipelines~\cite{he2024cameractrl, ren2025gen3c, videoworldsimulators2024, sun2025worldplay, yang2025layerpano3d}. However, a fundamental challenge remains: maintaining visual consistency over long temporal horizons. This challenge becomes particularly pronounced when the camera continuously moves and revisits previously observed regions, where the model is required to generate structurally consistent content across varying viewpoints. Furthermore, when editing operations are introduced into the video, such as global style changes or local object modifications, the model must preserve spatial coherence, and ensure that the edited results remain consistent with both past and future content. Achieving such consistency across time, viewpoints, and edits remains an open problem.

To address long-term consistency, existing methods~\cite{yu2025context, xiao2025worldmem, wu2025video, li2025vmem, zhang2025pretraining, zhang2025frame, wu2025corgi} have explored various memory-based strategies. Some recent studies~\cite{zhang2025pretraining, zhang2025frame, wu2025corgi} focus on temporal context modeling, where latent states or historical frames are stored to stabilize generation over time. Others~\cite{yu2025context, xiao2025worldmem, wu2025video, li2025vmem} leverage pose-conditioned feature retrieval, enabling the model to reuse previously observed visual information under similar viewpoints. These approaches improve temporal coherence and view consistency by incorporating historical information. However, they generally assume that past contexts remain valid throughout the generation process. This assumption breaks down in the presence of editing operations. When global edits (e.g., style transfer) or local edits (e.g., object replacement) occur, historical contexts may become partially or entirely outdated. As a result, models tend to rely on stale information, leading to semantic inconsistency, visual artifacts, or even reversion to pre-edit states.

In this work, we revisit the nature of spatial information in videos and identify a key insight: spatial context can be decomposed into two fundamentally different components—\textit{semantic appearance} and \textit{geometric structure}. These two components exhibit distinct temporal behaviors. Semantic appearance is highly dynamic and can change frequently due to editing operations, lighting variations, or style transformations. In contrast, geometric structure is typically stable over time or changes only locally. Existing methods entangle these two aspects within a unified representation, making the entire memory unreliable when semantics change. Consequently, even if the underlying geometry remains valid, it cannot be effectively reused due to its coupling with outdated appearance information. This observation motivates the need for a disentangled representation of spatial context, where semantic and geometric information can be independently updated and reused. Such a design allows the model to selectively refresh appearance while preserving stable structural knowledge, thereby enabling consistent generation across edits.

Based on this insight, we propose \textbf{PermaVid}, a novel framework for consistent video generation under editing operations. Our approach introduces a disentangled multi-modal context memory, where semantic appearance and geometric structure are modeled as separate but complementary memory representations. Specifically, we maintain an RGB-based memory to capture appearance information and a depth-based memory to encode geometry. To support flexible editing scenarios, we further design an edit-aware memory update and retrieval mechanism, which adheres to the division of video edits into global and local categories as presented in Ditto~\cite{bai2025ditto}. Global edits trigger semantic updates by invalidating outdated appearance memory while preserving geometric structure, whereas local edits selectively update only the affected regions. During generation, the model retrieves context from these memories in a spatially-aware manner and performs multi-modal feature fusion to guide novel view synthesis. This enables the model to generate videos that remain consistent across time, viewpoints, and editing operations. To enhance model learning capacity, we further leverage Unreal Engine with a built-in navigation agent to construct a long-form video dataset, termed \textbf{UE-Mem}. This dataset contains rich cross-scene revisiting trajectories, which enables our model to better learn the disentangled memory mechanism. Extensive experiments demonstrate that our method significantly outperforms existing approaches in maintaining both structural and semantic consistency under complex editing scenarios.


\section{Related Works}
\subsection{Video generation} 
Diffusion models~\cite{peebles2023scalable,ramesh2022hierarchical,rombach2022high} have become the dominant approach in video generation. Various efforts have been made to enhance the performance of video generation, including improvements in learning strategies~\cite{yang2024cogvideox,singer2022make}, data curation~\cite{qiu2023freenoise,li2024enhancing}, and prompt engineering~\cite{gao2025devil,long2025vista}. Video generation has progressed from U-Net–based models~\cite{wang2023modelscope,guo2023animatediff} to Transformer-based diffusion frameworks~\cite{videoworldsimulators2024,ma2024latte}, enabling realistic and temporally coherent videos. Recently, autoregressive approaches~\cite{chen2024diffusion, henschel2025streamingt2v} have been explored in video diffusion, reformulating generation from full-sequence denoising to a step-wise process.

\subsection{Camera-controlled Video Generation}
Camera-controlled video generation has emerged as an important direction toward controllable and interactive video synthesis. These methods introduce camera pose conditions or spatial constraints to guide generation along predefined trajectories, enabling scene exploration, viewpoint traversal, and revisiting~\cite{li2025hygamecraft,mao2025yume,sun2025worldplay}. By leveraging pose-conditioned generation or spatially-aware representations, these approaches partially alleviate geometric inconsistency and improve cross-view coherence. Despite these advances, existing methods still struggle to maintain global consistency over long temporal horizons, particularly under complex camera motions, viewpoint revisiting, or iterative editing operations. Camera-controlled approaches often rely on historical frames, latent states, or implicit memories to preserve cross-view coherence, but they are typically designed for static, unedited scenes. When editing introduces global or local edits, such methods may retrieve outdated context, leading to content reversion, inconsistent appearance, or structural misalignment.

\subsection{Memory-Augmented Video Models}
Memory mechanisms have been widely explored to enhance temporal coherence and controllability in video generation. Existing approaches introduce temporal memory, feature caching, or spatial retrieval mechanisms to retain historical information, such as compressed keyframes~\cite{zhang2025frame} or spatiotemporal context representations~\cite{yang2025cambrian,zhang2025pretraining}. In memory-based video generation, recent methods further employ pose-conditioned feature retrieval or external memory structures~\cite{sun2025worldplay,yu2025context,xiao2025worldmem,wu2025video,ren2025gen3c,li2025vmem} to preserve long-term consistency across time and viewpoints. However, most existing memory designs store past observations in a unified representation, where semantic appearance and geometric structure are implicitly entangled. Such coupled memories are sufficient when the scene state remains unchanged, but they become less flexible under editing operations. For instance, an appearance edit may invalidate the stored semantic content while leaving the underlying geometry still reusable, whereas a local object edit may only affect a spatially bounded region. Without separating these factors, existing methods lack a principled way to determine which memory entries should be preserved, updated, or discarded. This limitation motivates our proposed \textit{disentangled context memory}, which explicitly separates appearance-oriented RGB memory from geometry-oriented depth memory. This design enables edit-aware memory invalidation, selective regional updates, and reliable context reuse under editing operations.

\begin{figure*}[t]
	\centering
    \includegraphics[width=\linewidth]{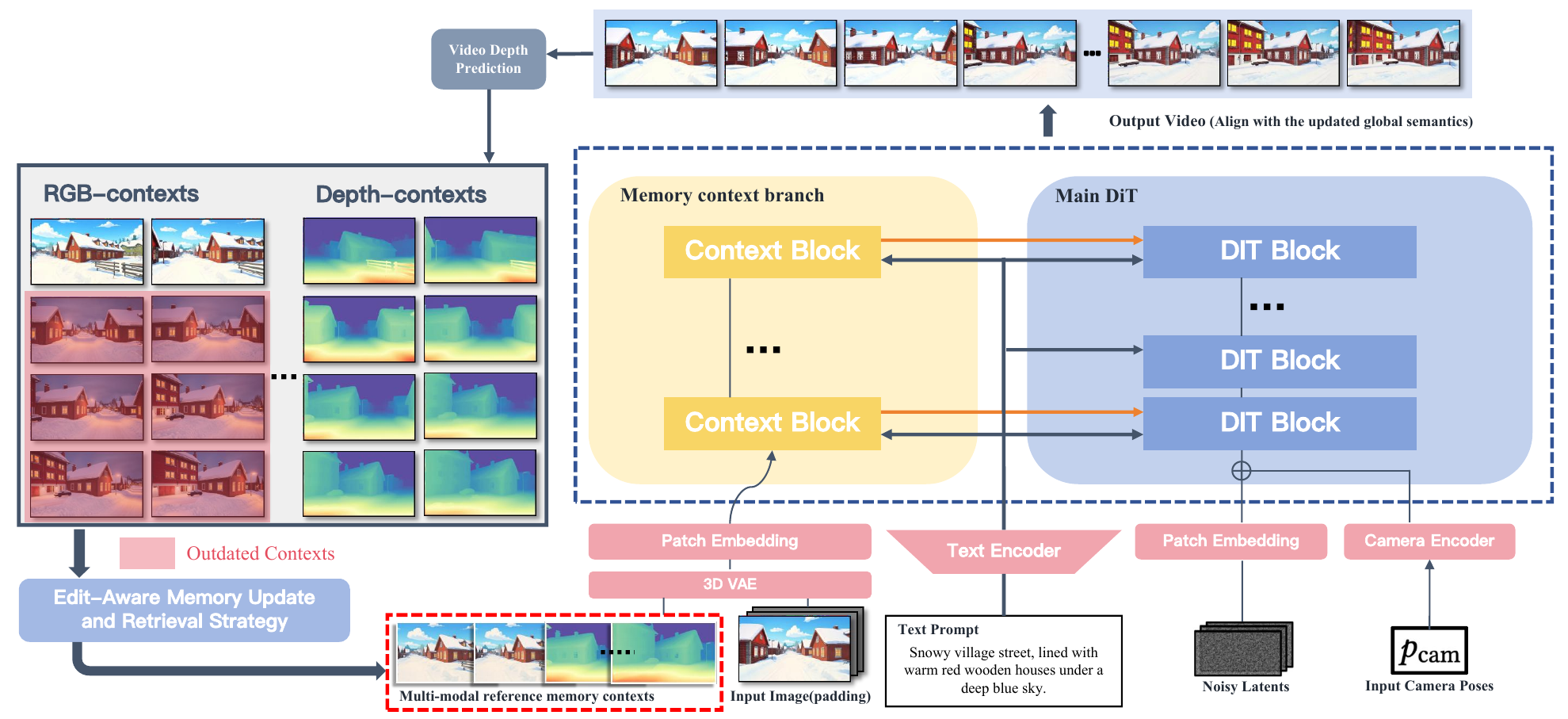}
        \setlength{\abovecaptionskip}{-3mm}
	\caption{\small
	\textbf{Overview of \textbf{PermaVid}. }PermaVid maintains a disentangled multi-modal context memory with an RGB bank for semantic appearance and a depth bank for geometric structure. Given target camera poses and editing operations, it updates and retrieves memory in an edit-aware manner, then fuses mixed-modality references to guide consistent video generation across time, viewpoints, and edits. 
	}
	\label{fig:pipeline}
    \vspace{-10pt}
\end{figure*}

\vspace{-5pt}
\section{Method}

Our goal is to achieve consistent video generation across edits, where edited content should persist in subsequent frames and remain coherent when changing viewpoints. This requires preserving the latest edited semantic appearance while maintaining reusable scene geometry over long temporal horizons. To this end, we revisit how spatial context is stored, updated, and retrieved in memory-based generation. Our key insight is that spatial context can be decomposed into \textbf{semantic appearance}, which may change frequently due to editing operations, and \textbf{geometric structure}, which is usually stable or changes only locally. Based on this observation, PermaVid introduces a disentangled multi-modal context memory (\cref{sec:method_memory}) that separates appearance from geometry, enabling selective memory updating, invalidation, and reuse across edits. During generation, PermaVid applies an edit-aware memory update and retrieval strategy (\cref{sec:method_strategy}) to select spatially relevant and valid references, and fuses the retrieved multi-modal context in a memory-guided video generation model (\cref{sec:method_simulator}). We further construct a synthetic training dataset with long revisiting trajectories and accurate camera poses to support learning this memory behavior (\cref{sec:method_dataset}). The pseudocode of the full pipeline is provided in the supplementary material.

\subsection{Disentangled Multi-modal Context Memory}
\label{sec:method_memory}

We decompose spatial context into two complementary factors: semantic appearance $A_t$ and geometric structure $G_t$. Semantic appearance includes visual attributes such as object identity, texture, color, illumination, and style, which may change after editing operations. Geometric structure describes scene layout, object shape, and spatial relationships, which are usually stable under global appearance edits and change only locally under object-level edits. This distinction motivates two separate memory banks:
\begin{equation}
\mathcal{M}^{\mathrm{rgb}}=\{(I_i^{\mathrm{rgb}}, \mathbf{p}_i, t_i, g_i)\}_{i=1}^{N},
\qquad
\mathcal{M}^{\mathrm{dep}}=\{(I_j^{\mathrm{dep}}, \mathbf{p}_j, t_j)\}_{j=1}^{M}.
\end{equation}
Here $I_i^{\mathrm{rgb}}$ and $I_j^{\mathrm{dep}}$ denote RGB and depth observations, $\mathbf{p}$ is the camera pose, $t$ is the timestamp, and $g_i$ records the global semantic version when an RGB memory unit is inserted.

The RGB memory provides high-fidelity appearance references for viewpoint-consistent synthesis. However, RGB observations naturally entangle appearance and geometry: the same image contains both semantic content and spatial layout. As a result, old RGB memory can become semantically outdated after an edit even when its underlying geometry is still useful. In contrast, the depth memory stores geometry-oriented context that is largely invariant to changes in texture, lighting, or style. By separating these modalities, PermaVid can refresh outdated appearance while preserving reusable structure, instead of discarding the entire historical context after every edit.

\subsection{Edit-aware Memory Update and Retrieval}
\label{sec:method_strategy}

PermaVid updates and retrieves memory according to the type and spatial extent of the editing operation. Following the video editing taxonomy in Ditto~\cite{bai2025ditto}, we consider two edit types: \textit{global edits} and \textit{local edits}. Global edits, such as style transfer, seasonal change, or lighting transformation, modify the semantic appearance of the whole scene while usually preserving its geometry. Local edits, such as object insertion, removal, or replacement, affect only a bounded region and may or may not change local geometry.

\paragraph{Memory update.}
For a global edit, all previously stored RGB contexts become semantically outdated, so PermaVid invalidates the RGB memory and advances the global semantic version $g^\ast$. The depth memory is retained because the scene geometry remains valid:
\begin{equation}
\mathcal{M}^{\mathrm{rgb}}\leftarrow\emptyset,\qquad
g^\ast\leftarrow g^\ast+1.
\end{equation}
For a local edit with affected region $\Omega_e$, PermaVid invalidates only memory units whose view footprint overlaps the edited region. Let $\Pi(\mathbf{p})$ denote the spatial footprint of a memory unit observed from pose $\mathbf{p}$. The RGB update is
\begin{equation}
\mathcal{M}^{\mathrm{rgb}}\leftarrow
\mathcal{M}^{\mathrm{rgb}}\setminus
\{m_i^{\mathrm{rgb}}\mid \Pi(\mathbf{p}_i)\cap\Omega_e\neq\emptyset\}.
\end{equation}
The depth memory follows the same local invalidation rule only when the edit changes geometry; otherwise, it is preserved. This selective update prevents stale edited content from being reused while keeping valid context in unaffected regions.

\paragraph{Memory retrieval.}
At generation time, the model retrieves references that are both spatially relevant to the target camera trajectory and valid under the current edited state. For each memory unit, we compute a trajectory-level overlap score
\begin{equation}
s_i=\max_{\mathbf{p}_q\in\mathbf{P}}\mathcal{L}(\mathbf{p}_i,\mathbf{p}_q),
\end{equation}
where $\mathbf{P}$ is the target camera trajectory and $\mathcal{L}$ measures normalized view-frustum overlap. RGB memory is retrieved only if it is spatially relevant and belongs to the current global semantic version:
\begin{equation}
R_c^{\mathrm{rgb}}=
\{m_i^{\mathrm{rgb}}\in\mathcal{M}^{\mathrm{rgb}}\mid s_i>\tau,\ g_i=g^\ast\}.
\end{equation}
Depth memory is retrieved based only on spatial relevance, since it is not invalidated by purely semantic edits:
\begin{equation}
R_c^{\mathrm{dep}}=
\{m_j^{\mathrm{dep}}\in\mathcal{M}^{\mathrm{dep}}\mid s_j>\tau\}.
\end{equation}
The final candidate set is $R_c^{\mathrm{rgb}}\cup R_c^{\mathrm{dep}}$. To keep the reference set compact, we greedily select at most $B$ memory units that cover the target trajectory while filtering out redundant views with high mutual overlap. This produces a spatially diverse mixed-modality reference set $\mathcal{R}^{\mathrm{mem}}$ for generation.

This update-and-retrieval design is the key mechanism that aligns memory with editing operations. Global edits refresh outdated appearance globally while preserving geometry; local edits only refresh affected regions; retrieval then avoids stale appearance memory while still reusing valid geometric context.

\subsection{Memory-guided Video Generation}
\label{sec:method_simulator}

Given the selected memory references $\mathcal{R}^{\mathrm{mem}}$, PermaVid uses a memory-guided video generation model built upon a diffusion Transformer (DiT). Target camera poses $p_{\mathrm{cam}}=[R,T]\in\mathbb{R}^{f\times(3\times4)}$ are encoded by a camera encoder and injected into the main DiT tokens, together with the text condition. To incorporate memory references, we add a dedicated memory context branch composed of distributed and cascaded Context Blocks duplicated from selected DiT layers.

The memory branch encodes RGB and depth references as mixed-modality conditions. Since retrieved memory frames are sparse observations sampled from different timestamps, each RGB or depth frame is independently encoded using a shared 3D VAE. The resulting memory tokens are concatenated with relative positional encoding within the memory set, avoiding reliance on absolute temporal indices and improving robustness to long temporal gaps. To better support heterogeneous conditions, the memory branch follows the VACE~\cite{jiang2025vace} context-branch design and is initialized from pretrained weights. The generation process can be summarized as
\begin{equation}
x_{t-1}=\epsilon_{\theta}\!\left(x_t,\,c_{\mathrm{text}},\,p_{\mathrm{cam}},\,\mathcal{R}^{\mathrm{mem}}\right),
\end{equation}
where $\mathcal{R}^{\mathrm{mem}}$ provides appearance and geometry references for consistent novel-view synthesis under edits.

\subsection{Dataset Construction}
\label{sec:method_dataset}

To support learning long-term memory behavior, we require long videos with revisiting trajectories and accurate camera poses. Existing public datasets either lack pose metadata or, like~\cite{li2025sekai, wang2025spatialvid}, mainly contain unidirectional camera motion without sufficient revisits. We therefore build an automatic data synthesis pipeline in Unreal Engine~5~\cite{UE_Spivey_2017,zhong2025unrealzoo}. Within each scene, a navigation agent combines two policies: a goal-driven navigator that explores distant regions and a pose tracker that induces local loops by following targets. Their combination naturally produces trajectories that revisit previously observed regions while maintaining broad scene coverage.

The agent uses a continuous 4-DoF control vector $\mathbf{u}=[v_x,v_y,\omega,g]$, where $v_x$ and $v_y$ denote lateral and forward velocity, $\omega$ is yaw rate, and $g$ is pitch angle. During recording, the agent is hidden, and a first-person camera rigidly attached to it captures long RGB and depth sequences with accurate 6-DoF camera poses. The resulting UE-Mem dataset contains 4k high-quality videos, each with 1000 frames, across 100 Unreal Engine scenes. We annotate each video caption using Qwen3-VL-7B~\cite{bai2025qwen3vl}.

\section{Experiments}

\subsection{Implementation Details} 
We implement our memory-guided video generation model based on the Wan2.1-14B~\cite{wan2025wan} architecture. During training, we employ variable-length video clips, with the number of frames randomly sampled between 25 and 81, with a fixed resolution of $480 \times 832$. To preserve the strong multi-modal feature perception capabilities in the memory context branch pretrained from VACE~\cite{jiang2025vace}, we keep this branch frozen and train only the parameters of the main DiT and camera encoder. The training process is conducted in two stages. In the first stage, we train the model on the public short-video dataset SpatialVid~\cite{wang2025spatialvid} with camera pose annotations to activate its camera-guided video generation capability. Training is performed on 32 NVIDIA H200 GPUs for 10k steps. Specifically, in this stage, the input image is padded to the target video length and fed into the memory context branch without any additional reference images. In the second stage, we train the model on our constructed long-video dataset UE-Mem, which features revisiting trajectories, to further learn memory-guided generation from mixed multi-modal reference contexts. This stage is conducted for 8k training steps. Among the training samples, 40\% use pure RGB reference contexts, 40\% use pure depth reference contexts, and the remaining 20\% employ mixed-modality contexts.
At inference time, we set the memory context size to 10, indicating that references are retrieved from 10 distinct views, with each view providing information in either the RGB or depth modality. At each autoregressive iteration, the last frame of the previously generated video chunk is used as the input image. The generated video is then fed into a depth predictor~\cite{chen2025vda} to obtain the corresponding depth contexts, which are subsequently incorporated into our multi-modal context memory, as illustrated in Figure~\ref{fig:pipeline}.


\subsection{Metrics and Baselines.}

To evaluate consistent video generation across edits, we compare our approach against existing state-of-the-art methods, including HY-Worldplay~\cite{sun2025worldplay}, HY-Gamecraft~\cite{li2025hygamecraft}, Matrix-Game-2.0~\cite{he2025matrix}, and VMem~\cite{li2025vmem}. We construct a benchmark test set of 200 images sourced from free websites and AI-generated images, covering both realistic and stylized scenes as well as indoor and outdoor environments. The evaluation set includes random and complex camera trajectories with viewpoint revisiting, making it suitable for testing long-term cross-view consistency after edits. For each case, all methods are evaluated with the same predefined action scripts, including the input camera trajectory and the prompt and timing of the editing operation. To ensure a controlled comparison, we use Qwen-Image~\cite{wu2025qwenimage} to apply the edit during streaming generation for all methods. We evaluate two settings following the global/local edit taxonomy. 1) For the \textbf{long-term consistency after local edits}, we measure view recall consistency and visual quality. View recall consistency is measured using PSNR, SSIM, and LPIPS on paired RGB frames captured after the edit at the same camera location, within video sequences generated along forward and reversed camera trajectories. Visual quality is measured using VBench~\cite{huang2024vbench}. 2) For the \textbf{long-term consistency after global edits}, we decompose view recall consistency into structural consistency, measured by PSNR, SSIM, and LPIPS on paired depth frames, and semantic consistency, evaluated by CLIP-Vid similarity~\cite{clip}, since global edits change semantic appearance while preserving geometry.

\vspace{-5pt}

\subsection{Qualitative Comparison}

\begin{figure*}[t]
	\centering
    \includegraphics[width=\linewidth]{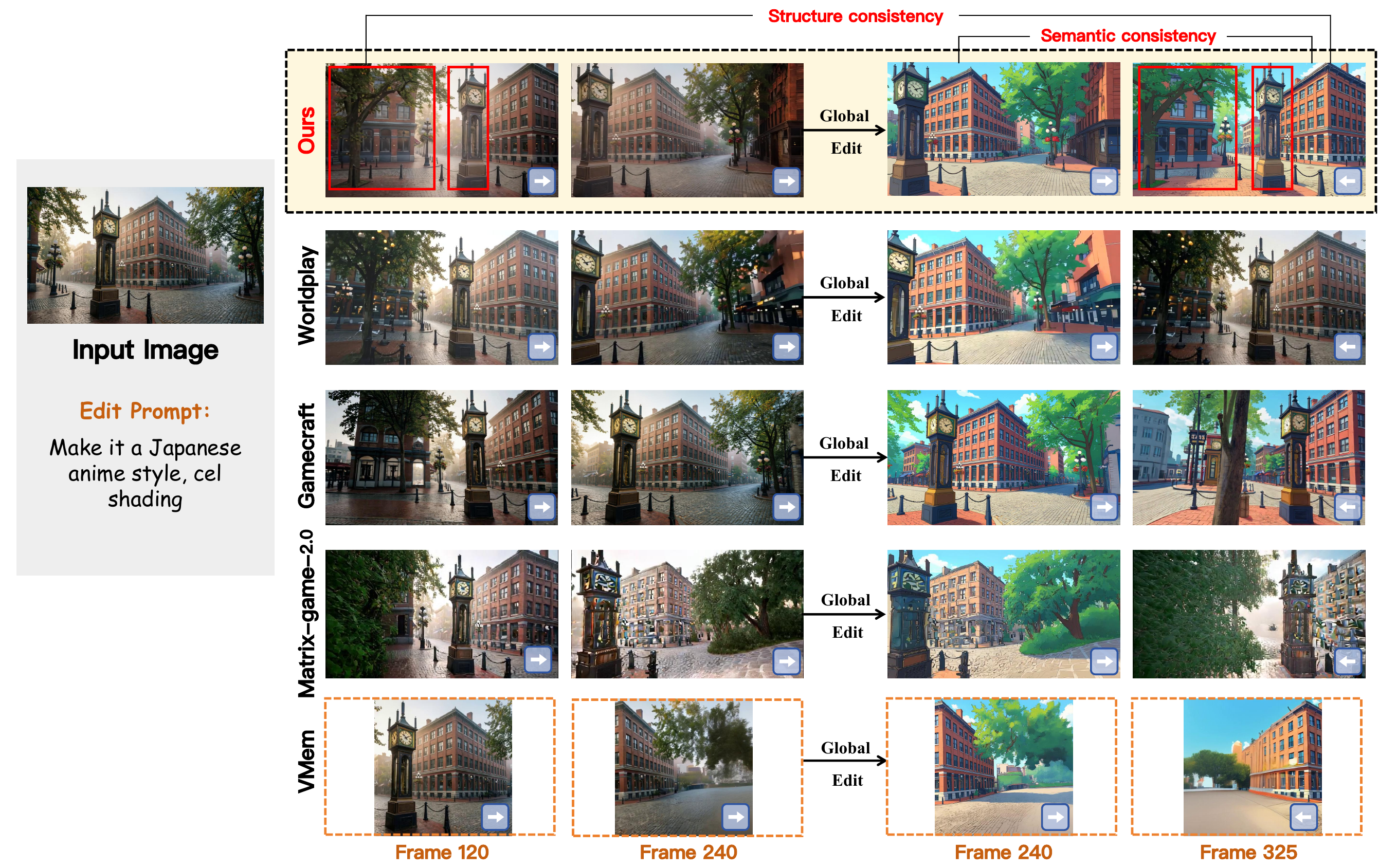}
    \vspace{-2pt}
        \setlength{\abovecaptionskip}{-1mm}
	\caption{\small
	\textbf{Qualitative comparison under global edits.} Under a global edit (e.g., style transformation), our method maintains stable geometric structure while consistently propagating the edited semantic appearance across time and viewpoints.
	}
    \vspace{-12pt}
	\label{fig:qualitative_global}
\end{figure*}

\begin{table}[t]
\centering
\small
\caption{Quantitative comparison across edits. We achieve the best overall performance under both \textbf{global} and \textbf{local} edits, demonstrating strong long-term consistency.}
\label{tab:quantitative_global_local}
\vspace{-6pt}

\resizebox{\linewidth}{!}{
\begin{tabular}{l c c c c c c c c c c c c c}
\toprule
\multirow{3}{*}{Methods}
& \multicolumn{7}{c}{\textbf{Global Edits}}
& \textbar
& \multicolumn{5}{c}{\textbf{Local Edits}} \\

\cmidrule(lr){2-8}\cmidrule(lr){10-14}

& \multicolumn{3}{c}{Structure Consistency}
& \textbar
& \multicolumn{1}{c}{Semantics consistency}
& \textbar
& \multicolumn{1}{c}{Visual Quality}
& \textbar
& \multicolumn{3}{c}{View Recall Consistency}
& \textbar
& \multicolumn{1}{c}{Visual Quality} \\

\cmidrule(lr){2-4}\cmidrule(lr){6-6}\cmidrule(lr){8-8}
\cmidrule(lr){10-12}\cmidrule(lr){14-14}

& PSNR$\uparrow$
& SSIM$\uparrow$
& LPIPS$\downarrow$
& \textbar
& CLIP-Vid$\uparrow$
& \textbar
& VBench-Avg$\uparrow$
& \textbar
& PSNR$\uparrow$
& SSIM$\uparrow$
& LPIPS$\downarrow$
& \textbar
& VBench-Avg$\uparrow$ \\
\midrule

HY-Worldplay
& \underline{22.34} & \underline{0.8674} & \underline{0.2339}
& \textbar
& 19.96
& \textbar
& 0.8488
& \textbar
& \underline{21.922} & \underline{0.8523} & \underline{0.2470}
& \textbar
& 0.8320 \\

HY-Gamecraft
& 18.83 & 0.8194 & 0.2951
& \textbar
& \underline{26.17}
& \textbar
& \textbf{0.8691}
& \textbar
& 15.492 & 0.4388 & 0.4351
& \textbar
& \underline{0.8377} \\

Matrix-Game-2.0
& 16.40 & 0.7065 & 0.4054
& \textbar
& 21.86
& \textbar
& 0.8417
& \textbar
& 14.408 & 0.3394 & 0.3465
& \textbar
& 0.8086 \\

VMem
& 14.92 & 0.6621 & 0.4386
& \textbar
& 18.74
& \textbar
& 0.8264
& \textbar
& 16.207 & 0.7018 & 0.3216
& \textbar
& 0.7893 \\


\midrule
\textbf{Ours}
& \textbf{22.84} & \textbf{0.8703} & \textbf{0.2102}
& \textbar
& \textbf{27.87}
& \textbar
& \underline{0.8565}
& \textbar
& \textbf{22.332} & \textbf{0.8622} & \textbf{0.2369}
& \textbar
& \textbf{0.8544} \\

\bottomrule
\end{tabular}
}

\end{table}

\vspace{-5pt}
\subsubsection{Long-term consistency after global edits}
We conduct qualitative comparisons with other state-of-the-art methods, focusing on two key criteria: structural consistency and semantic consistency. This evaluation is motivated by the fact that global edits alter the overall semantic appearance, while the underlying geometric structure should remain stable. As shown in Figure~\ref{fig:qualitative_global}, after applying a global edit at frame 240, the camera revisits previously observed viewpoints along the trajectory. Our method preserves structural consistency under revisiting views, as evidenced by the stable spatial layout and intact geometric structures (e.g., building placements and clock tower geometry). At the semantic level, our model updates the scene to reflect the latest edited appearance, enabling coherent style propagation across different viewpoints and over time. In contrast, other methods exhibit clear memory degradation. HY-WorldPlay preserves geometric consistency under reversed viewpoints but fails to update semantics, resulting in outdated styles after the global edit. HY-Gamecraft and Matrix-Game-2.0 fail to maintain consistency in both structural and semantic aspects. VMem performs better than purely short-context baselines in some consistency-related metrics, but still lags behind our method due to its limited ability to preserve sharp structure while updating edited semantics. This suggests that simply reusing historical memory is insufficient for stable long-term propagation without explicit disentanglement between geometry and appearance.

\vspace{-10pt}

\subsubsection{Long-term consistency after local edits}

\begin{figure*}[t]
	\centering
    \includegraphics[width=\linewidth]{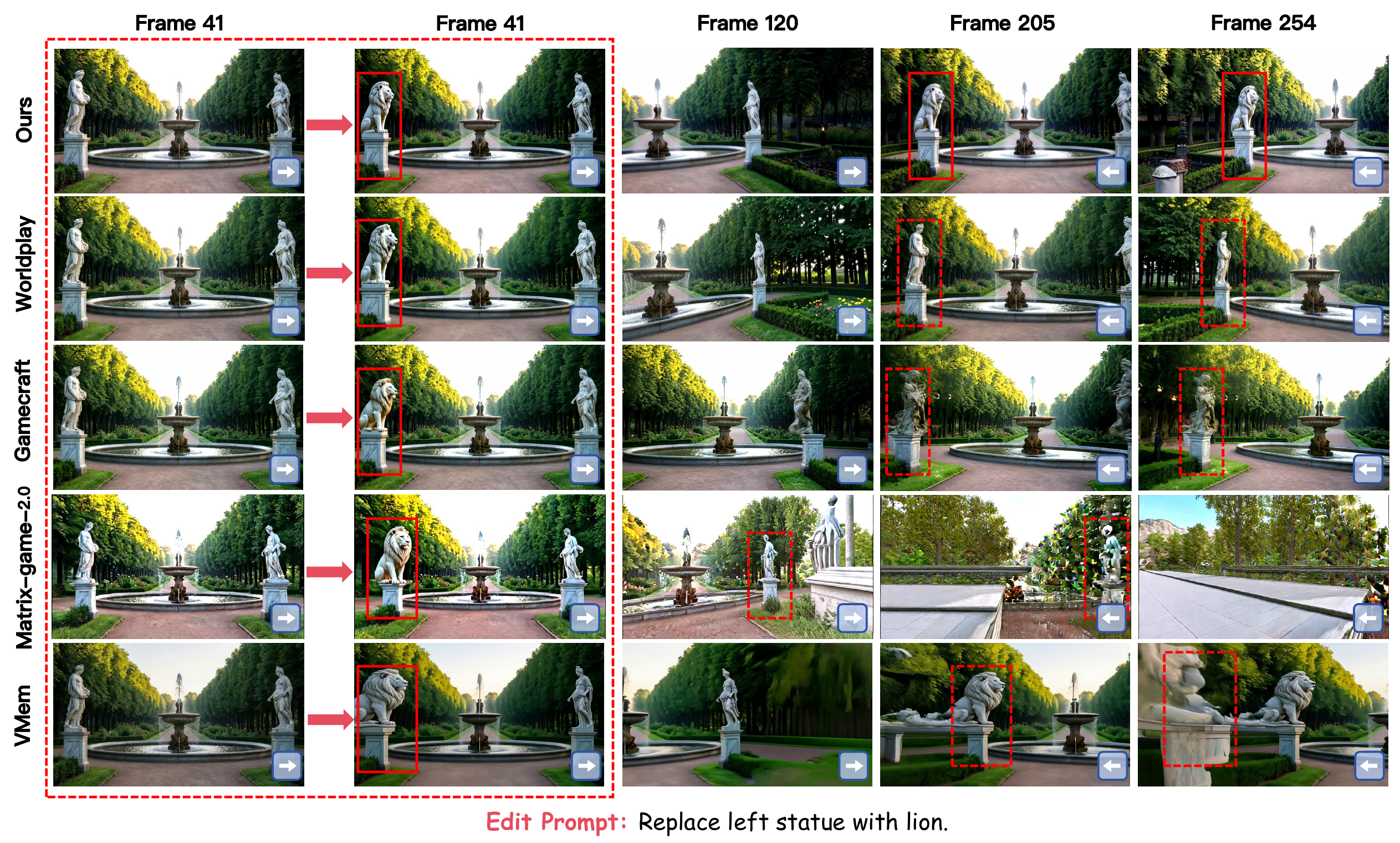}

        \setlength{\abovecaptionskip}{-1mm}
	\caption{\small
	\textbf{Qualitative comparison under local edits.} Under a local edit, our method consistently recalls the edited region during revisiting while preserving the surrounding geometric structure.
	}
    
    \vspace{-8pt}
	\label{fig:qualitative_local}
\end{figure*}

\vspace{-4pt}

We evaluate view recall consistency under local edits by examining whether the edited content is correctly recalled when the camera revisits the same viewpoints. As shown in Figure~\ref{fig:qualitative_local}, after a local edit is applied to a specific region, our method consistently recalls the post-edit content during revisiting, preserving both structural integrity and updated local semantics.
In contrast, HY-WorldPlay fails to properly perceive and retain local edits: although it maintains view consistency during revisits, the recalled content corresponds to the pre-edit state, indicating insufficient edit awareness. HY-Gamecraft and Matrix-Game-2.0 exhibit severe inconsistency, as their limited short-term context (e.g., chunk-level KV cache) is insufficient to support reliable view recall over time. VMem can recall the edited object under revisiting, but the recalled content is blurry and spatially unstable. This indicates that memory retrieval alone does not guarantee accurate local edit preservation when appearance and geometry are not explicitly coordinated.

\vspace{-5pt}

\subsection{Quantitative Comparison}
\subsubsection{Long-term consistency after global edits}

We quantitatively evaluate long-term consistency after global edits by decomposing view recall consistency into structural and semantic consistency. As shown in Table~\ref{tab:quantitative_global_local}, our method achieves the best performance in PSNR, SSIM, and LPIPS, indicating strong preservation of geometric structure under global semantic edits. It also significantly outperforms all baselines in semantic consistency (CLIP-Vid), reflecting effective propagation of the edited appearance. In contrast, existing methods either preserve structure but fail to update semantics (e.g., HY-WorldPlay) or degrade in both aspects due to the absence of persistent edit-aware memory (e.g., VMem, HY-GameCraft and Matrix-Game-2.0).

\subsubsection{Long-term consistency after local edits}

We quantitatively evaluate long-term consistency after local edits in terms of view recall consistency and visual quality. As shown in Table~\ref{tab:quantitative_global_local}, our method achieves the best results in PSNR, SSIM, and LPIPS, indicating accurate recall of edited local content when revisiting the same viewpoints. Moreover, our approach also delivers the best visual quality measured by VBench-Avg. In contrast, baseline methods show obvious degradation in view recall consistency, reflecting their inability to reliably retain and recall local edits over time.

\vspace{-8pt}
\subsection{Ablation Study}

\begin{figure}[t]
    \centering
    \includegraphics[width=0.92\linewidth]{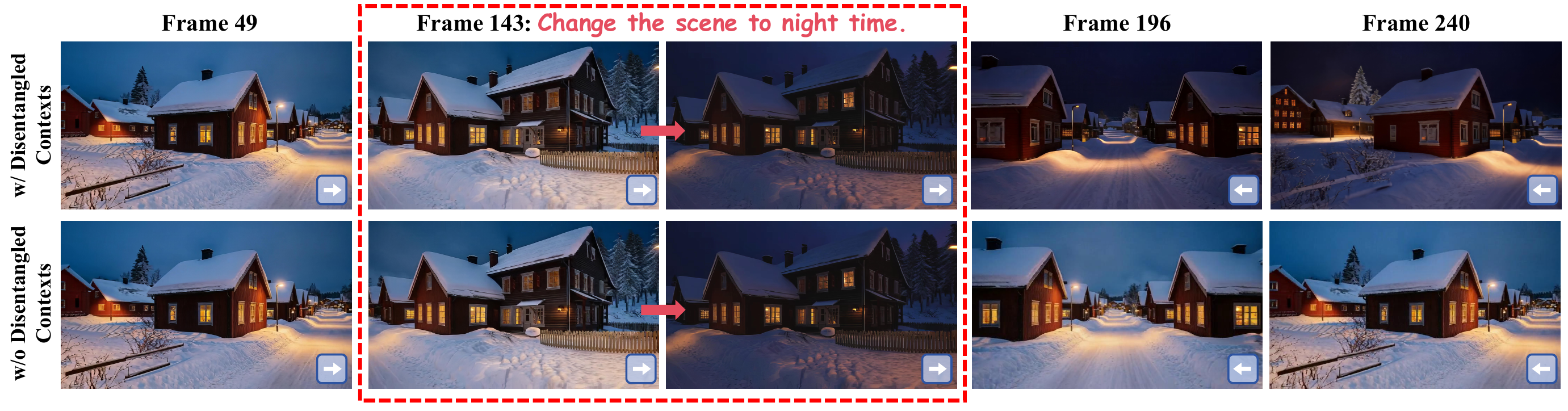}
    \setlength{\abovecaptionskip}{0.5mm}
    \setlength{\belowcaptionskip}{-1mm}
    \caption{\small
    \textbf{Ablation study on disentangled context memory.} With disentangled context memory, the model consistently propagates updated global semantics after the edit while preserving stable geometry, whereas entangled RGB contexts reuse outdated semantics, leading to degraded global semantic consistency over time.}
    \label{fig:ablation_mmem}
\end{figure}

We ablate the proposed disentangled context memory by comparing two settings: \textit{w/ Disentangled Contexts} and \textit{w/o Disentangled Contexts}. w/ Disentangled Contexts uses the proposed multi-modal memory design, where retrieved reference contexts can be RGB-only, depth-only, or mixed-modality depending on the edit-aware retrieval strategy. In this setting, semantic appearance and geometric structure are handled separately, allowing the model to reuse geometry while selectively updating outdated appearance information after edits. By contrast, w/o Disentangled Contexts stores all historical contexts only in RGB form and directly reuses them through the same camera-based retrieval index, without separating appearance from geometry or filtering out semantically outdated contexts after edits. As shown in Figure~\ref{fig:ablation_mmem}, after a global day-to-night edit at Frame~143, \textit{w/ Disentangled Contexts} consistently propagates the updated semantics across subsequent frames and revisited views while preserving stable geometric structure. In contrast, \textit{w/o Disentangled Contexts} repeatedly retrieves outdated RGB contexts, which prevents it from reflecting the latest global edit and leads to progressive semantic inconsistency over time.

\vspace{-8pt}

\subsection{Memory Overhead Analysis}
\label{sec:memory_overhead_analysis}

\begin{figure}[t]
    \centering
    \includegraphics[width=\linewidth]{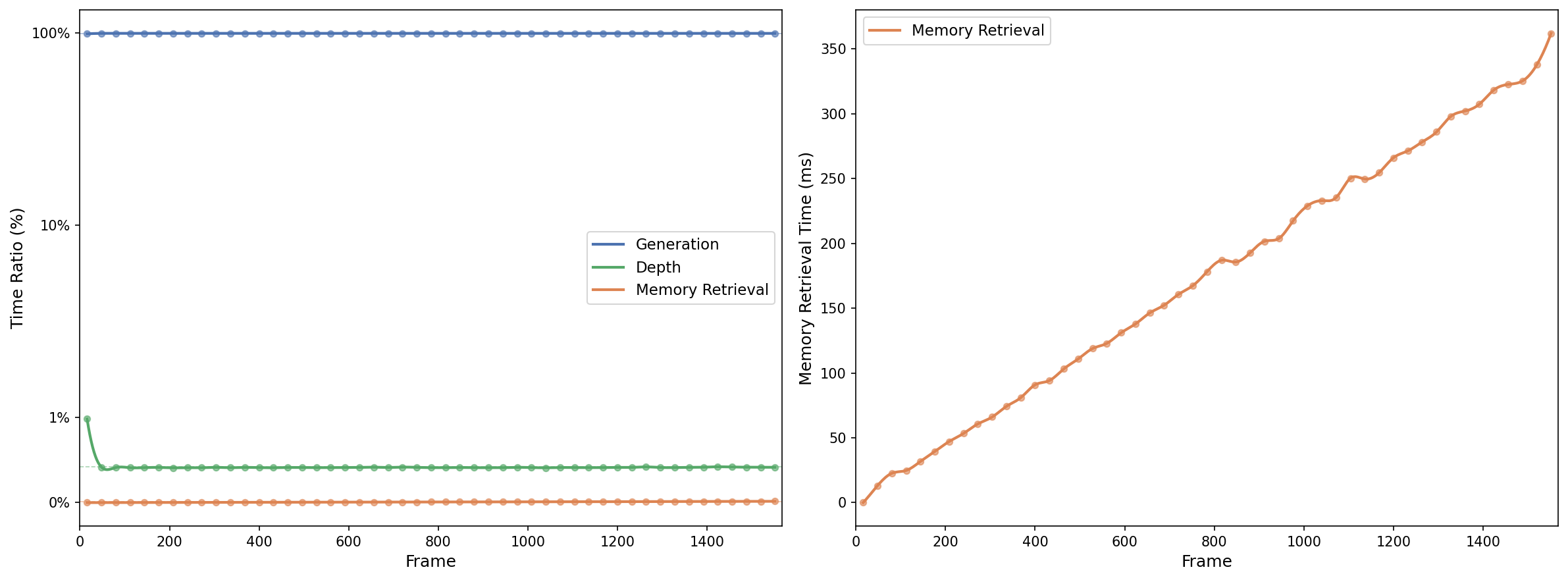}
    \caption{
Memory overhead profiling during long-duration generation. Left: component time ratios show that inference is dominated by the video generation backbone, while depth prediction and memory retrieval contribute negligibly. Right: memory retrieval time gradually increases as historical contexts accumulate, but remains at the millisecond level, indicating only marginal runtime overhead.
    }
    \label{fig:memory_overhead_live}
\end{figure}

We further analyze the computational overhead introduced by the proposed memory mechanism during long-horizon generation. To quantify this overhead, we profile both the relative runtime ratio of each component and the absolute retrieval time throughout a long generation sequence with a large-loop camera trajectory. Figure~\ref{fig:memory_overhead_live} shows that the overall runtime is dominated by the video generation backbone. As illustrated in the left plot, the generation module consistently accounts for nearly all inference time, while depth estimation and memory retrieval remain negligible across the entire sequence. This result indicates that the additional operations required for memory maintenance and retrieval contribute only a very small fraction of the total computational cost. The right plot reports the absolute retrieval time as the generation proceeds. As expected, retrieval becomes gradually slower as more historical contexts accumulate in memory. Nevertheless, the growth remains mild, and the retrieval cost stays at the millisecond level even near the end of the sequence, reaching only a few hundred milliseconds. This overhead is negligible relative to the diffusion-based video generation process and does not constitute a practical bottleneck. Overall, these results show that PermaVid scales efficiently to long-video generation by maintaining and reusing disentangled multi-modal contexts with only marginal runtime overhead.

\vspace{-5pt}
\section{Conclusion}
In this work, we study how to preserve long-term consistency in video generation across editing operations. To address the limitations of existing memory mechanisms, we propose a multi-modal context memory that disentangles spatial context into semantic appearance and geometric structure, together with an edit-aware memory update and retrieval strategy. By selectively updating and reusing memory across modalities, our approach keeps retrieved context aligned with the latest edited state while preserving reusable geometry. We further build a memory-guided video generation model that performs multi-modal feature fusion over modality-asymmetric reference contexts, enabling consistent generation across time, viewpoints, and edits. Extensive experiments under both global and local edits demonstrate the effectiveness of our approach, showing strong semantic and structural consistency and clear advantages over state-of-the-art methods.


{
\bibliographystyle{unsrt}
\bibliography{main}

}




\begin{figure}[h]
    \centering
    \includegraphics[width=\textwidth, height=\textheight, keepaspectratio]{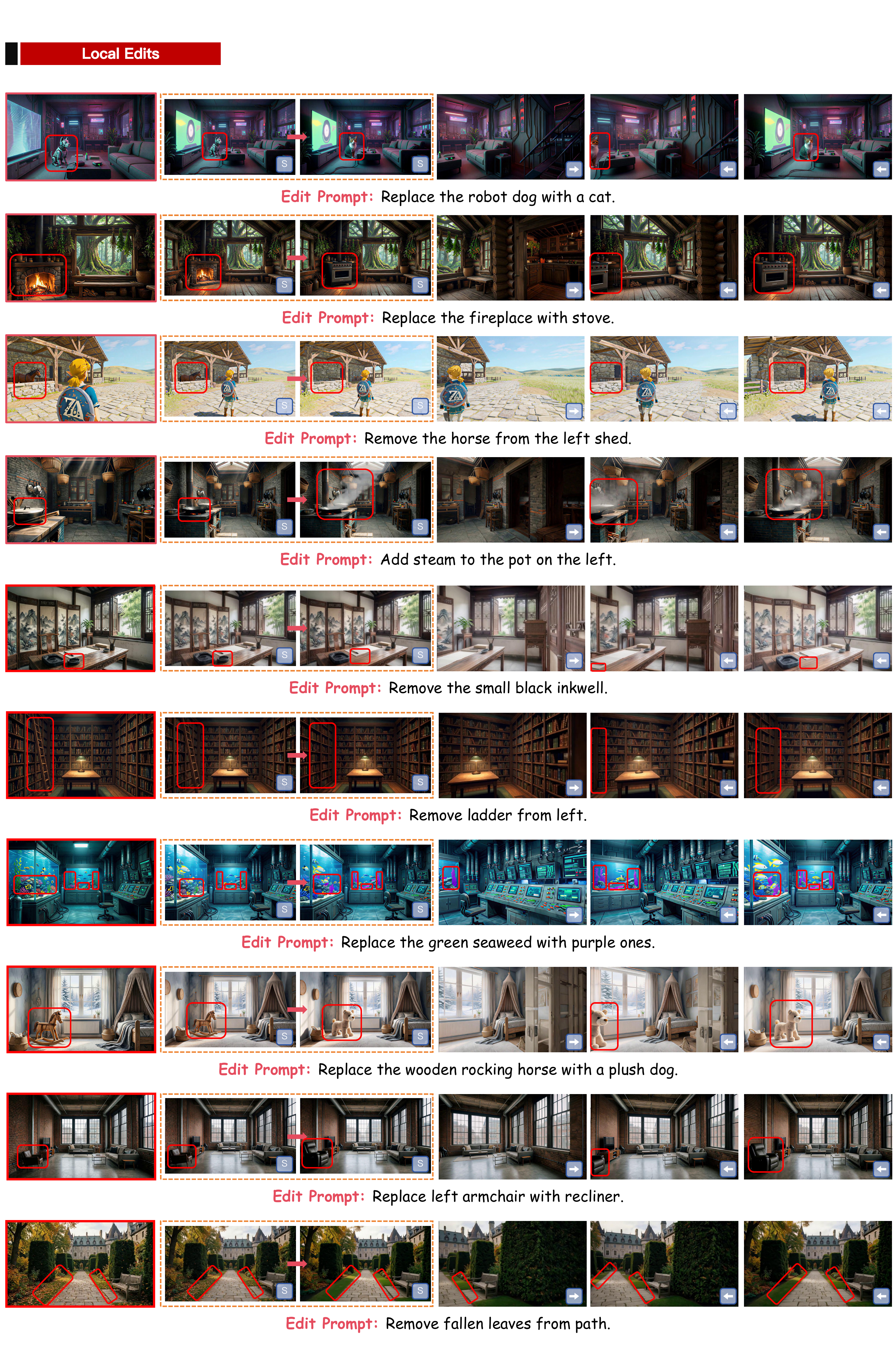}
    \caption{Additional results under local edits, showing localized semantic updates with preserved scene geometry and consistent behavior under revisited views.}
    \label{fig:more_results_local}
\end{figure}

\begin{figure}[h]
    \centering
    \includegraphics[width=\textwidth, height=\textheight, keepaspectratio]{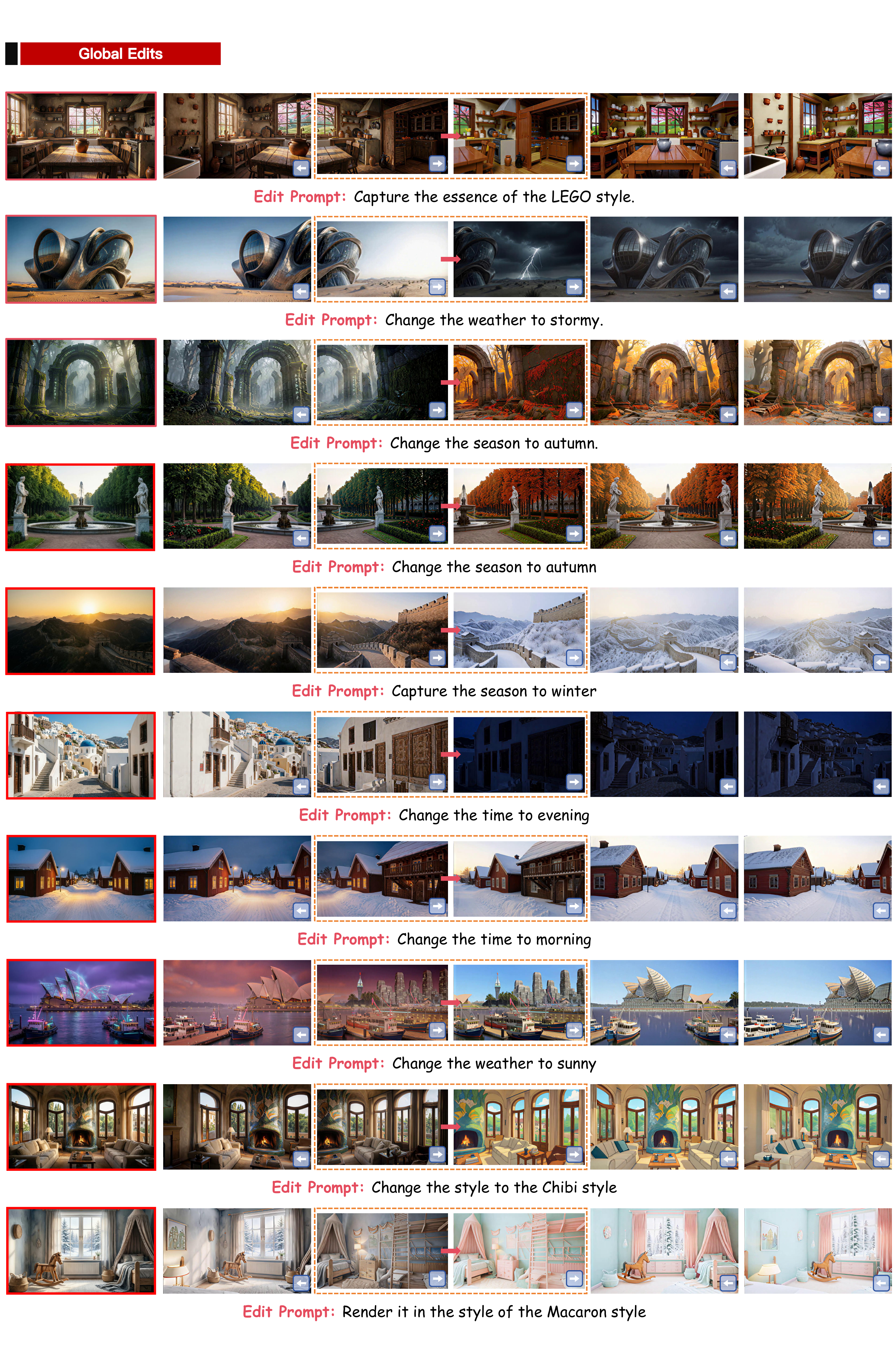}
    \caption{Additional results under global edits, showing coherent propagation of global semantic attributes (e.g., style, illumination, and season) across revisited views with stable scene geometry.}
    \label{fig:more_results_global}
\end{figure}

\end{document}